\documentclass{article}
\usepackage{amssymb}
\usepackage{graphicx}
\usepackage{caption}
\usepackage{amsmath}
\usepackage[final]{corl_2025} % Uncomment for the camera-ready ``final'' version.
\title{Affordance-Based Disambiguation of Surgical Instructions for Collaborative Robot-Assisted Surgery}

\author{
  Ana Davila \\
  Nagoya University, 
  Japan\\
  \And
  Jacinto Colan  \\
  Nagoya University, 
  Japan\\
  \And
  Yasuhisa Hasegawa \\
  Nagoya University, 
  Japan\\
}

\begin{document}
\maketitle

%===============================================================================

\begin{abstract}
Effective human-robot collaboration in surgery is affected by the inherent ambiguity of verbal communication. This paper presents a framework for a robotic surgical assistant that interprets and disambiguates verbal instructions from a surgeon by grounding them in the visual context of the operating field. The system employs a two-level affordance-based reasoning process that first analyzes the surgical scene using a multimodal vision-language model and then reasons about the instruction using a knowledge base of tool capabilities. To ensure patient safety, a dual-set conformal prediction method is used to provide a statistically rigorous confidence measure for robot decisions, allowing it to identify and flag ambiguous commands. We evaluated our framework on a curated dataset of ambiguous surgical requests from cholecystectomy videos, demonstrating a general disambiguation rate of 60\% and presenting a method for safer human-robot interaction in the operating room.
\end{abstract}

\keywords{ambiguity, LLMs, affordance, surgical robots}

%===============================================================================

\section{Introduction}
	
In collaborative robot-assisted surgery, verbal commands from surgeons are often pragmatically ambiguous \cite{davila25beyond}. An instruction such as ``cut" is clear in vocabulary but underspecified in context, lacking the specific tool and target \cite{sutkin2024speech}. This ambiguity in the core components of a request (tool, action, object) can lead to incorrect actions and risk patient safety. Current systems often use rigid command structures that limit natural interaction  \cite{liu2024review, colan25hrcollaboration}. We propose a framework that resolves this ambiguity by integrating visual perception with knowledge-based reasoning. Our method uses the concept of tool affordances, the actions that a tool can perform on an object, to infer the specific executable command from the visual context \cite{tagliabue2022deliberation}. To ensure safety, we incorporate a conformal prediction layer to provide a quantifiable measure of confidence, allowing the robot to know when it should request clarification rather than making a high-risk guess \cite{fontana2023conformal}.

%===============================================================================
\section{Proposed Framework}

Our framework processes a surgeon's verbal instruction and the corresponding endoscopic video feed to produce a validated, executable command for a robotic assistant \cite{davila24voice}. The pipeline consists of three main stages: visual grounding, affordance reasoning, and conformal prediction for safety validation, as depicted in Figure \ref{fig:framework}.

% \begin{figure}[h]
%   \centering
%   \includegraphics[width=0.5\linewidth]{./crest25_workshop_davila-figure-1.png}
%   \caption{Overview of the proposed framework, showing the flow from surgeon instruction to robot execution. The system processes speech and endoscopic images through two levels of analysis before using conformal prediction to decide if a command is safely executable.}
%   \label{fig:framework}
% \end{figure}

\subsection{Level 1: Visual Grounding Analysis}

The first stage grounds the linguistic instruction in the visual context of the surgical scene.
\begin{enumerate}
    \item \textbf{Object Recognition:} An object detection model, fine-tuned on surgical data using an adaptive transfer learning strategy, processes the live endoscope image to identify and locate surgical tools and anatomical structures.
    \item \textbf{Vision Expert:} The outputs of the object detector (class labels) are formatted as a structured prompt for a multi-modal Vision Expert model (VLM). This model analyzes visual evidence to generate a structured JSON description of the scene, detailing the presence and state of each identified tool and object (e.g. ``grasper is present and holding tissue"). 
\end{enumerate}

\subsection{Level 2: Tool-augmented Affordance Reasoning}

This stage uses a Reasoning Expert, implemented with a Large Language Model, to disambiguate the instruction.
\begin{enumerate}
    \item \textbf{Surgical Affordance Knowledge Base:} We define a knowledge base that contains information about tool affordances. This can be represented as a set of logical predicates, `CanPerform(tool\_type, action, object\_type)', which returns true if the specified tool can perform the action on the object (e.g., `CanPerform(`cutter', `cut', `tissue')' is true).
    \item \textbf{Chain-of-Thought Reasoning:} The Reasoning Expert uses a chain-of-thought process, augmented with function calls (tools), to validate the instruction against the visual scene description and the affordance knowledge base. It first checks if the required tool is in the scene and then verifies if the requested interaction is valid. This process generates intermediate reasoning steps and concludes with a set of binary ambiguity scores: $S_{tool\_missing}$, $S_{action\_invalid}$, and $S_{target\_unclear}$.
\end{enumerate}

\subsection{Safety through Dual-set Conformal Prediction}

To provide a safety guarantee, we use Dual-set Conformal Prediction, which offers a robust confidence metric without making assumptions about the data distribution \cite{davila25llmbased}.
\begin{enumerate}
    \item \textbf{Nonconformity Score:} We define a nonconformity score function, $\mathcal{A}(x, y)$, which measures how ``unusual" a sample $x$ is with respect to a class $y \in \{\text{Ambiguous, Non-Ambiguous}\}$. The score is derived from the ambiguity flags generated by the Reasoning Expert.
    \item \textbf{Calibration:} The system is calibrated offline using a dataset $C$ containing 120 instruction-image pairs, split into two sets: $C_{Amb}$ and $C_{NonAmb}$. For each sample $i$ in these sets, we compute its nonconformity score $\alpha_i = \mathcal{A}(x_i, y_i)$.
    \item \textbf{Prediction:} For a new test sample $x_{test}$, we hypothesize that it belongs to each class and calculate two separate p-values:
    \begin{equation}
        p_{Amb} = \frac{|\{i \in C_{Amb} : \alpha_i \geq \alpha_{test}\}| + 1}{|C_{Amb}| + 1}
    \end{equation}
    \begin{equation}
        p_{NonAmb} = \frac{|\{i \in C_{NonAmb} : \alpha_i \geq \alpha_{test}\}| + 1}{|C_{NonAmb}| + 1}
    \end{equation}
    These p-values represent the credibility of the sample belonging to the 'Ambiguous' and 'Non-Ambiguous' classes, respectively. Based on a predefined significance level $\alpha$ (e.g., 0.1), we apply decision rules: if $p_{NonAmb} > \alpha$ and $p_{Amb} \leq \alpha$, the instruction is deemed 'Executable'. Other outcomes lead to 'Ambiguous' or 'Uncertain' classifications, prompting feedback from the surgeon.
\end{enumerate}

\begin{figure*}[t]
    \centering
    % Left minipage for the large framework diagram
    \begin{minipage}[c]{0.54\textwidth}
        \centering
        \includegraphics[width=\linewidth]{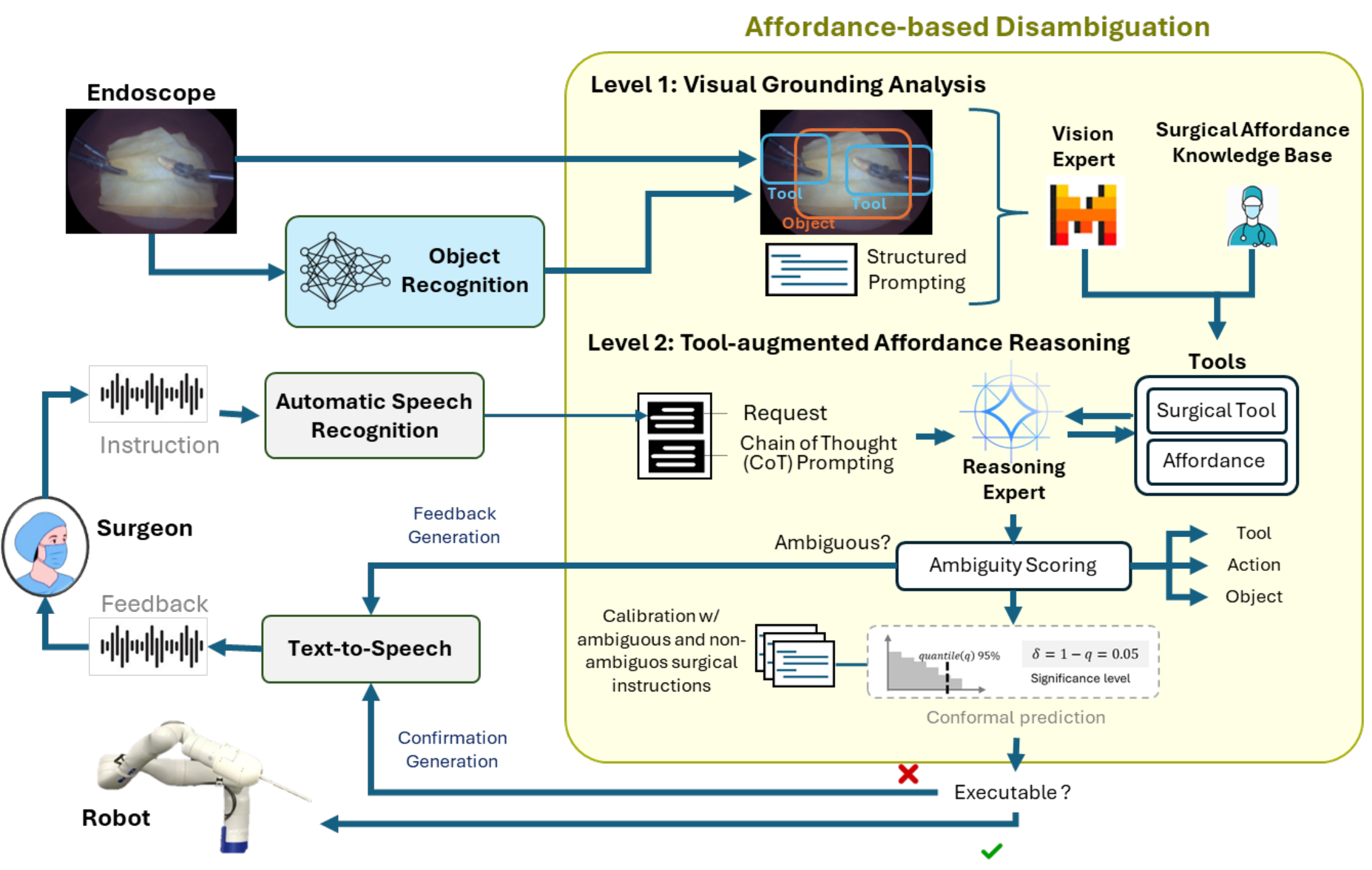}
        \caption{Overview of the proposed framework, showing the flow from surgeon instruction to robot execution. The system processes speech and endoscopic images through two levels of analysis, then uses conformal prediction to ensure the command is safely executable.}
        \label{fig:framework}
    \end{minipage}\hfill
    % Right minipage for the stacked plot and table
    \begin{minipage}[c]{0.42\textwidth}
        \centering
        % Top item: p-value plot
        \includegraphics[width=\linewidth]{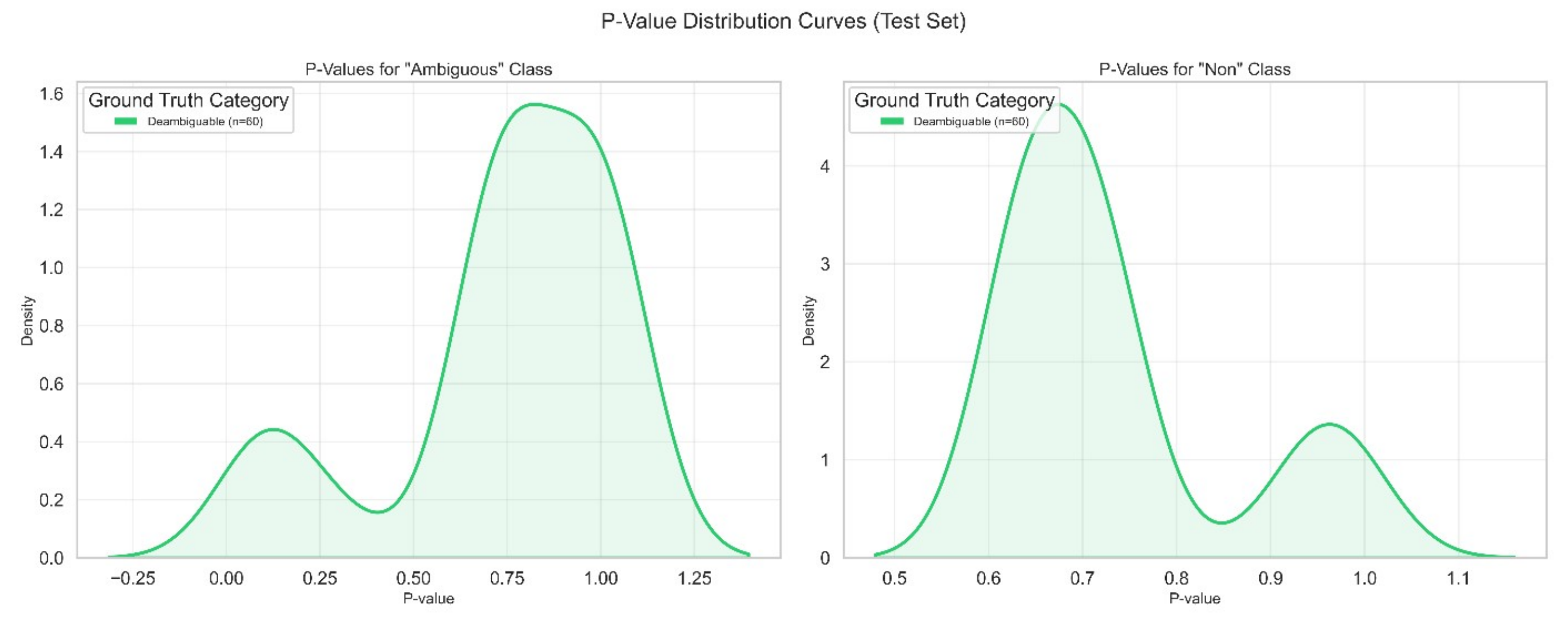}
        \caption{P-value distribution for ambiguous and non-ambiguous classes.}
        \label{fig:p_values}
        \vspace{1em} % Vertical space between the plot and the table
        % Bottom item: Results table
        \captionof{table}{Disambiguation Outcomes for 60 Ambiguous Requests}
        \label{tab:results}
        \centering
        \resizebox{\linewidth}{!}{%
        \begin{tabular}{lcc}
            \hline
            \textbf{Outcome Category} & \textbf{Count} & \textbf{Percentage} \\
            \hline
            Successfully Disambiguated & 35 & 58.3\% \\
            Failed Disambiguation (Still Ambiguous) & 9  & 15.0\% \\
            Flagged as Uncertain & 16 & 26.7\% \\
            \hline
            \textbf{Total} & \textbf{60} & \textbf{100\%} \\
            \hline
        \end{tabular}%
        }
    \end{minipage}
\end{figure*}

%===============================================================================

\section{Experimental Evaluation}
\label{sec:result}

\subsection{Dataset and Setup}

We curated a dataset of 180 surgical scenes from the Cholec80 cholecystectomy video dataset \cite{nwoye23cholectriplet}. Each scene was paired with a verbal request manually labeled into one of three categories: \textbf{Unambiguous:} All three components (tool, action, object) are clearly identified; \textbf{Deambiguable:} Contains ambiguity in one or more components, but can be resolved using visual context; \textbf{Truly Ambiguous:} Ambiguity cannot be resolved from visual input alone.
% \begin{itemize}
%     \item \textbf{Unambiguous:} All three components (tool, action, object) are clearly identified.
%     \item \textbf{Deambiguable:} Contains ambiguity in one or more components, but can be resolved using visual context.
%     \item \textbf{Truly Ambiguous:} Ambiguity cannot be resolved from visual input alone.
% \end{itemize}
The evaluation focused on a test set containing the 60 deambiguable surgical requests, specifically targeting ambiguities related to tool presence and action-affordance mismatches. The calibration set for conformal prediction consisted of the other 120 samples.

\subsection{Results and Analysis}

Our method achieved an overall disambiguation rate of approximately 60\% across all ambiguity types. For ambiguities related to tool presence, the performance was higher at 75\%. The outcomes for the test set of 60 initially ambiguous requests are detailed in Table \ref{tab:results}. The system successfully disambiguated 35 of the requests (58.3\%), correctly inferring the missing information. In 9 cases (15.0\%), the system failed to resolve the ambiguity. Importantly, the conformal prediction framework identified its own lack of confidence in 16 cases (26.7\%), correctly flagging them as 'Uncertain' and thus avoiding potentially unsafe actions.

The effectiveness of the conformal prediction framework is visualized in Figure \ref{fig:p_values}. The kernel density estimate shows a clear separation between the p-value distributions for the ambiguous and non-ambiguous classes from the calibration set. This separation indicates that the nonconformity score is effective at distinguishing between the two classes, enabling the system to make confident predictions.

%===============================================================================

\section{Conclusion}
\label{sec:conclusion}

This work presented a framework that resolves pragmatic ambiguity in verbal surgical instructions by combining affordance-based reasoning with conformal prediction for safety. Our hybrid approach allows a robotic assistant to interpret underspecified commands and quantify its own uncertainty, establishing a methodology for safer human-robot communication. The 58.3\% success rate in disambiguating requests is a promising initial result. Future work will focus on improving accuracy with larger datasets, optimizing for real-time performance, and enabling the system to learn new tool affordances from video data.

%===============================================================================

\clearpage
% The acknowledgments are automatically included only in the final and preprint versions of the paper.
% \acknowledgments{This work was supported in part by the Japan Science and Technology Agency (JST) CREST under Grant JPMJCR20D5, and in part by the Japan Society for the Promotion of Science (JSPS) Grants-in-Aid for Scientific Research (KAKENHI) under Grant 22K14221.}

%===============================================================================

\bibliography{biblio}  % .bib

\end{document}